\documentclass[conference]{IEEEtran}
\usepackage{cite}
\usepackage{amsmath,amssymb,amsfonts}
\usepackage{algorithmic}
\usepackage{graphicx}
\usepackage{textcomp}

\usepackage[english]{babel}
\usepackage[T1]{fontenc}
\usepackage{epsfig}
\IEEEoverridecommandlockouts
\usepackage{tabu}
\usepackage{float}
\usepackage{amsmath}
\usepackage{amssymb}
\usepackage{amsthm}
\usepackage[keeplastbox]{flushend}
\usepackage{cite}
\usepackage{subfigure}
\usepackage{epstopdf}
\usepackage{multirow}
\usepackage{caption}
\usepackage[table]{xcolor}
\usepackage{mathtools}
\usepackage{xfrac}
\usepackage{units}
\usepackage[hyphens]{url}
\usepackage{array}
\newcommand{\beql}[1]{\begin{equation}\label{#1}}
\newcommand{\eeq}{\end{equation}}

\newcommand{\be}{\begin{equation}}
\newcommand{\ee}{\end{equation}}
\newcommand{\ba}{\begin{array}}
\newcommand{\ea}{\end{array}}

\ifCLASSINFOpdf
\else
\fi
\begin{document}
%

\title{Impact of $L_1$ Batch Normalization on Analog Noise Resistant Property of  Deep Learning Models}
%
%
%

\author{Omobayode~Fagbohungbe,~\IEEEmembership{Student~Member,~IEEE,}
        ~Lijun~Qian,~\IEEEmembership{Senior~Member,~IEEE}
\thanks{The authors are with the CREDIT Center and the Department
of Electrical and Computer Engineering, Prairie View A\&M University, Texas A\&M University System, Prairie View, 
TX 77446, USA. Corresponding author: Omobayode Fagbohungbe e-mail: ofagbohungbe@pvamu.edu}}
\maketitle


\begin{abstract}
Analog hardware has become a popular choice for machine learning on resource-constrained devices recently due to its fast execution and energy efficiency. However, the inherent presence of noise in analog hardware and the negative impact of the noise on deployed deep neural network (DNN) models limit their usage. The degradation in performance due to the noise calls for the novel design of DNN models that have excellent noise-resistant property, leveraging the properties of the fundamental building block of DNN models. In this work, the use of $L_1$ or $TopK$ BatchNorm type, a fundamental DNN model building block, in designing DNN models with excellent noise-resistant property is proposed. Specifically, a systematic study has been carried out by training DNN models with $L_1$/$TopK$ BatchNorm type, and the performance is compared with DNN models with $L_2$ BatchNorm types. The resulting model noise-resistant property is tested by injecting additive noise to the model weights and evaluating the new model inference accuracy due to the noise. The results show that $L_1$ and $TopK$ BatchNorm type has excellent noise-resistant property, and there is no sacrifice in performance due to the change in the BatchNorm type from $L_2$ to $L_1$/$TopK$ BatchNorm type.

\end{abstract}


\section{Introduction}
\label{sec:Introduction}

Large scale datasets, high-performance hardware, algorithmic and architectural techniques, and more sophisticated optimization methods have continued to play a critical role in the unprecedented success of deep neural networks (DNN) in notoriously complex and highly challenging real-life cognitive applications such as computer vision, speech recognition, machine translation, autonomous driving, and anomaly detection~\cite{resnet,Burrgeo2021,Onasami2022UnderwaterAC,Yang2022AJE,Fagbohungbe2021EfficientPP}. They achieve this by enabling the design and training of deep and large state-of-the-art models with complex architectures and remarkable performance.

However, these large models require an order of millions of multiply-accumulate operations, which are fundamentally computational intensive operations~\cite{Kariyappa}. These data-intensive operations and a large number of model parameters due to the model size mean that these models would require large memory and memory bandwidth in order to achieve reasonable performance~\cite{mixedsignal,Li,joshi,charan}. As a result, these deep models cannot be deployed on resource-constrained edge computing devices with limited computing resources and power budget~\cite{NoisyNN,mixedsignal} such as battery-powered mobile and internet of things (IoT) devices.

In order to resolve these issues, model compressing methods such as pruning~\cite{han2015deep,luo2017thinet} and knowledge distillation~\cite{hinton2015distilling} have been introduced to design lighter models. Although a significant reduction in model size has been achieved using these methods, the resulting models still require a decent amount of computing, memory, and power resources. These reasons have led to the introduction of specialized energy-efficient computing hardware such as TPU~\cite{tpu} and low power GPUs~\cite{gpu} for deep learning inference called digital accelerators. The von-Neumann architecture nature of these hardware means that the performance improvement is limited due to the intrinsic bottleneck between memory and processor known as the memory wall~\cite{Chen}. The memory wall causes the hardware to suffer from high energy consumption and high latency due to the substantial movement of data between the memory and the processor~\cite{xiao,joshi}.

On the other hand, analog accelerators, which are based on non-von-Neumann architecture, offer exciting opportunities for significantly more efficient and faster DNN accelerators~\cite{Kariyappa}. They offer analog computation with in-memory computing, leveraging on non-volatile memory (NVM) crossbar arrays to store and perform compute in the analog domain~\cite{Kariyappa,Chang}. With the weights encoded and stored in the memory arrays, the matrix-vector multiplication can be done in a single step using the Kirchhoff current law, unlike multiple steps required in digital hardware~\cite{Kornilios2020}. This form of computation offers huge performance improvement as it affords massive parallelism using a dense array of millions of memory devices performing computation~\cite{Chutengarmibm}. Furthermore, the combination of compute and storage in a single unit significantly reduces data movement and communication, which is the cause of high energy consumption and high latency in digital accelerators~\cite{Kornilios2020,Chutengarmibm}.

Despite the advantages of analog accelerators, it should be noted that computation in them is inherently noisy, and only limited precision can be achieved. The conductance noise due to inaccuracies in programming analog conductance (programming noise) and inherent non-idealities of NVM devices such as white noise, thermal noise, quantization noise, variability, and conductance drift all contribute to the noise in analog accelerators~\cite{mixedsignal,NoisyNN,Kariyappa}. The presence of noise leads to degradation in the inference performance of DNN models even with the known noise-resistant property of DNN~\cite{NoisyNN,joshi,Fagbohungbe2020BenchmarkingIP,Merolla2016DeepNN}. Hence, there is a need to explore methods for the novel design of DNN models that have excellent noise-resistant property, leveraging the properties of the fundamental building block of DNN models.

To design DNN models that have significant noise-resistant property, the impact of the various model building block and hyperparameters on the model noise-resistant property of DNN models must be investigated. One of the most famous building blocks for designing DNN models is Batch Normalization (BatchNorm). BatchNorm popularity can be attributed to its ability to stabilize the distribution of input over a mini-batch into a network during training by augmenting the network with additional layers that set the mean and variance of the distribution of each activation to zero and one, respectively~\cite{Santurkar2018HowDB}, as defined in equations (\ref{mean})-(\ref{me2}), where $\gamma$ and $\beta$ are the trainable parameters.
\begin{equation}
\label{mean}
{\mu_{B}}\leftarrow\frac{1}{m} {\sum\limits_{i=1}^{m} x_i}
\end{equation}
\begin{equation}
\label{me}
\hat{x_{i}}\leftarrow\frac{x_i - \mu_{B}}{\sigma_{B}}
\end{equation}
\begin{equation}
\label{me2}
{y_{i}}\leftarrow \gamma\hat{x_i} + \beta \equiv BN(x_i)
\end{equation}

The BatchNorm layer, usually placed before the non-linear activation of the previous layer, also preserves model expressivity by ensuring that the output of its layer is also scaled and shifted based on the model trainable parameters~\cite{Santurkar2018HowDB}. BatchNorm also improves regularization due to the randomness it introduces via its mini-batch statistics and reduces the difficulty of annealing learning rate and initializing parameters by enabling bigger learning rate~\cite{Bjorck2018UnderstandingBN,Ioffe2017BatchRT,Ioffe2015BatchNA,Wu2019L1B}.

\begin{table*}
\centering
\caption{The details of the DNN models and dataset used in the experiments.}
\label{table:details}
\begin{tabular}{|*{6}{c|} }
\hline
Model Name & Dataset & Number of Classes & Model Input Dimension & \# Images per class\\
\hline\hline
ResNet & [CIFAR10, CIFAR100] & [10,100] & 32*32*3 & [6000,600] \\
\hline
ResNext & [CIFAR10, CIFAR100] & [10,100] & 32*32*3 & [6000,600]\\
\hline
DenseNet & [CIFAR10, CIFAR100] & [10,100] & 32*32*3 & [6000,600] \\
\hline
\end{tabular}
\end{table*}

Batch Normalization can be classified into three types based on how the variance of the mini-batch is calculated as the mean of the mini-batch is calculated the same way. The three types are $L_2$ BatchNorm, $L_1$ BatchNorm, and $TopK$ BatchNorm. The $L_2$ BatchNorm is the most popular type of BatchNorm, and the variance of $L_2$ normalization is calculated according to equation (\ref{var}) which is the average squared deviation from the mean. $L_1$ BatchNorm calculates its variance by using the average absolute deviation from the mean as stated in equation (\ref{var1}). Instead of using all data in the mini-batch as it obtained in $L_2$ BatchNorm and $L_1$ BatchNorm, the $TopK$ BatchNorm uses the top K maximum absolute deviation from the mean to calculate its variance, in this case, $k=10$ as given in equation (\ref{var2}).
\begin{equation}
\label{var}
{\sigma_{B}}\leftarrow \sqrt{\frac{1}{m} {\sum\limits_{i=1}^{m} {(x_i-\mu_{B})}^2} - \epsilon}
\end{equation}
\begin{equation}
\label{var1}
{\sigma_{B}}\leftarrow {\frac{1}{m} {\sum\limits_{i=1}^{m} {|x_i-\mu_{B}|}}}
\end{equation}
\begin{equation}
\label{var2}
{\sigma_{B}}\leftarrow {\frac{1}{10} {\sum\limits_{i=1}^{10} {|x_i-\mu_{B}|}}}
\end{equation}

The pervasiveness of BatchNorm requires that its impact on the noise-resistant property of DNN models need to be investigated. The work in ~\cite{BatchNorm2020} studied the impact of $L_2$ BatchNorm on the noise-resistant property of DNN by comparing the performance between DNN models with BatchNorm and DNN models without BatchNorm. The paper discovered that $L_2$ BatchNorm \emph{negatively} impacts DNN noise-resistant property significantly. It attributes this effect to the ability of $L_2$ BatchNorm to limit the upper bound of the power of the noise inherently present in the training process. The authors also theorized that models with $L_2$ BatchNorm layer have gradient loss with a low range of values in every direction and that $L_2$ distance between the two consecutive instances of the loss gradient is significantly lower for a model with BatchNorm than the models without BatchNorm. The presence of noise in the training process has been used to improve the noise-resistant property of DNN models in the past. Hence, the paper showed that $L_2$ BatchNorm reduces the power of the inherent noise present during training, which now reduces the noise-resistant property of the resulting model. It concluded that models for analog accelerators are best designed without the $L_2$ BatchNorm layer. If model convergence is not possible without BatchNorm, the DNN model should be designed with the minimum number of BatchNorm layers needed to achieve convergence during model training instead of the current practice of having a BatchNorm after each layer.

The training of the DNN model without the $L_2$ BatchNorm layer is non-trivial as the model convergence might be difficult to achieve without it as $L_2$ BatchNorm stabilizes model training. Furthermore, it enables the use of a larger learning rate and also makes training deep and large state-of-the-art models possible by helping reduce internal covariate shift (ICS)~\cite{Ioffe2015BatchNA}. In cases where model convergence is possible without it, the size of the model and the complexity of the task are very limited. Hence, there is a need to modify/replace the current $L_2$ BatchNorm algorithm with another algorithm that preserves the current advantages of the $L_2$ BatchNorm algorithm and also improve the model noise-resistant property. The use of $L_1$ and $TopK$ BatchNorm types for the design and training of DNN model for analog accelerators is proposed to solve these problems. These BatchNorm types are selected as the variance calculation in the $L_2$, and $L_1$/TopK  BatchNorm types are strongly correlated to preserve the stabilizing property of $L_2$ BatchNorm.

The noise-resistant property of DNN models is strongly influenced by the noise inherently present in the training process or externally injected into the model weight, model weight gradient, or model input during the training process~\cite{joshi,NoisyNN}. The inherent noise present in the training process introduces some variability into the gradient of the weight and it is mathematically defined by the equation (\ref{eequ4}) as $\alpha\nabla L_D(x)$ represents the gradient of the weight and $\frac{\alpha}{B}{\sum_{i \in B} ({\nabla L_i(x) - \nabla L_D(x)})} $ represents the error term where $\alpha$ is the learning rate, $\nabla$ is the gradient, $ L_D(x)$ is the loss function over all the dataset, $ L_i(x)$ is the loss function for a single data point $i$, $\nabla L_{SGD}(x)$ is the estimated true loss gradient, and $B$ is the Batch size.
\begin{equation}
\label{eequ4}
\alpha \nabla_{SGD}(x)= \alpha \nabla L_D(x) + \frac{\alpha}{B}{\sum_{i \in B} ({\nabla L_i(x) - \nabla L_D(x)})}
\end{equation}

\begin{equation}
\label{eequ7}
E[\|{{\alpha\nabla L_B(x) - \alpha\nabla L_{SGD}(x)}}\|^2] < \frac{\alpha^2}{|B|}{C}
\end{equation}

\begin{equation}
\label{eequ6}
C=E[\|{{\nabla L_i(x) - \nabla L_D(x)}}\|^2]
\end{equation}

The error/noise term has a mean value of zero, and the upper bound of the power of the noise is defined by equations (\ref{eequ7}) and (\ref{eequ6}) where $L$ represents the loss function. The noise of higher power introduces more noise into the weight gradient, producing a model with excellent noise-resistant property. The poor noise-resistant of DNN models with $L_2$ BatchNorm is because the normalization process influences the size of the power of the noise as the value of $C$ is influenced by BatchNorm type. Since the value of the $\sigma_{B}$ for $L_1$ and $TopK$ BatchNorm type is lower than for $L_2$ BatchNorm, the power of the inherent noise present for model with $L_1$/$TopK$ BatchNorm is higher than the same models with $L_2$ BatchNorm. Hence, DNN models with $L_1$/$TopK$ BatchNorm has excellent noise-resistant property than DNN models with $L_2$ BatchNorm. Furthermore, the relation between the loss gradient of the weight of a model with BatchNorm and those without BatchNorm is shown in equation (\ref{eequ9})~\cite{Santurkar2018HowDB} where $L$ and $\hat{L}$ are the loss of the model with and without BatchNorm, respectively, $\sigma_j$ is the standard deviation of the BatchNorm, $\gamma$ is the BatchNorm layer trainable parameter, where ${y_j}$ and $\hat{y_j}$ are the output of the model with and without BatchNorm, respectively,$\nabla$ is the gradient, and $m$ is the batch size.
\begin{equation}
\label{eequ9}
\| \nabla _{y_j}\hat{L}\|^2 \leq \frac{\gamma^2}{\sigma_j^2} \Big( \| \nabla _{y_j}L\|^2] - \frac{1}{m} {\big \langle{1,\nabla _{y_j}L}\big \rangle}^2 - \frac{1}{m} {\big \langle{\nabla _{y_j}L, \hat{y_j}}\big \rangle}^2 \Big)
\end{equation}

The equation also shows that the loss gradient of the weight is also influenced by the size of the $\sigma_j$. A low $\sigma_j$ value means that the gradient value is higher than when $\sigma_j$ is high. The presence of noise can be explained by the changing nature of the $\sigma_j$ value from one epoch to another. Hence, with a lower $\sigma_j$ value for $L_1$/$TopK$ BatchNorm type compared with the $L_2$ BatchNorm type, the value of the loss gradient of DNN model and power of the noise present is higher for the DNN model with $L_1$/$TopK$ BatchNorm type than the same DNN model with $L_2$ BatchNorm type.

Hence, a detailed experimental study to verify the suitability of the $L_1$ and $TopK$ BatchNorm types for the design and training of DNN models for analog accelerators is performed in this work. This study involves training various DNN model architecture with $L_2$, $L_1$, and $TopK$ BatchNorm types on the Image classification task. After that, additive Gaussian noise is injected into all the weights of the resulting models, and the impact of the noise on the inference accuracy of the model is measured. A comparison is then performed between the $L_2$, $L_1$, and $TopK$ BatchNorm types variant of the models using the normalized inference accuracy, which better reflects the rate of degradation. The contributions of this research work are:
\begin{enumerate}
\item Proposed and detailed a way to design and train DNN models with great noise-resistant property for analog accelerators;
\item Established the effect of $L_1$ and $TopK$ BatchNorm types on the noise-resistant property of DNN models;
\item Provided justification for the use of $L_1$ BatchNorm type for the design of noise-resistant DNN models;
\item Provided a mathematical explanation on why the effect of $L_1$ and $TopK$ BatchNorm types on the noise-resistant property of a DNN model differ from the effect of $L_2$ BatchNorm type.

\end{enumerate}

The remainder of this paper is organized as follows: The detailed proposed methodology for this work is discussed in Section~\ref{sec:methods}. Results and analysis are presented in Section~\ref{sec:results}. Further discussions and related works are reviewed in Section~\ref{sec:discussion} and Section~\ref{sec:conclusion} concludes the paper.

\section{Proposed Methodology}
\label{sec:methods}
The discussion in this section focuses on the details of the experiment carried out to study the effect of $L_1$ and $TopK$ batch normalization layer on the inherent resistance characteristics of deep learning models to analog noise. The details discussed are the dataset, model design, model training, model inferencing, and software and hardware used.

\subsection{Dataset}
\label{subsec:ExpSetup}

The details of the CIFAR10 and CIFAR100 datasets used in this paper are stated in Table~\ref{table:details}. Each dataset contains 60,000 colored images of dimension $32\times32\times3$, further divided into 50,000 training images and 10,000 testing images. They are designed, compiled, and labeled by the Canadian Institute for Advanced Research out of the 80 million tiny images datasets. The images in the CIFAR10 dataset can be grouped into ten mutually exclusive classes with no semantic overlap. However, CIFAR100 dataset images can be grouped into 100 non-mutually exclusive classes with some form of semantic overlap as the dataset contain just 20 superclasses.

\subsection{Model Design}
\label{subsec:ExpSetup}
The models used in this paper can be generally grouped into ResNet~\cite{resnet}, ResNext~\cite{xie2017aggregated}, and DenseNet~\cite{huang2018densely} models as stated in Table~\ref{table:details}. The choice of the models is influenced by their suitability for image classification tasks and also to verify the impact of the various BatchNorm types on noise-resistant properties of diverse deep learning models. The various models are trained until convergence is achieved by minimizing the prediction error and maximizing the model accuracy using the default initialization methods in PyTorch, categorical cross-entropy as the cost function, stochastic gradient descent as the optimizer. Data augmentation Is also performed to prevent overfitting and maximize model performance.

\begin{table*}
\centering
\caption{Comparison of the baseline inference accuracy of ResNet, ResNext and DenseNet models with various BatchNorm types when tested with CIFAR10 and CIFAR100 dataset. The performance metric is the model classification accuracy.}
\label{tab:result_baseline}
\begin{tabular}{|c|c|c|c|c|c|c|c|c|c|c|c|c|c|c|}
\hline
\multirow{3}{4em}{} & \multicolumn{6}{|c|}{Dataset}\\ \hline

& \multicolumn{3}{|c|}{CIFAR10}& \multicolumn{3}{|c|}{CIFAR100}\\ \hline

{Model Name} & {$L_2$ BN} & {$L_1$ BN} & {TopK BN}& {$L_2$ BN} & {$L_1$ BN} & {TopK BN}\\ \hline

{Resnet18} & {90.87\%} & {90.35\%}& {89.50\%}& {65.79\%} & {65.39\%}& {64.73\%}\\ \hline

{Resnet36} & {92.32\%} & {91.74\%}& {91.69\%}& {68.99\%} & {68.81\%}& {68.91\%}\\ \hline

{Resnet44} & {92.87\%} & {92.53\%}& {92.46\%}& {70.28\%} & {69.44\%}& {69.50\%}\\ \hline

{Resnet56} & {92.70\%} & {92.67\%}& {92.67\%}& {70.48\%} & {70.14\%}& {70.58\%}\\ \hline

{ResNext18} & {91.53\%} & {91.40\%}& {90.90\%}& {68.20\%} & {67.61\%}& {67.88\%}\\ \hline

{Resnext56} & {93.50\%} & {93.07\%}& {93.02\%}& {72.25\%} & {71.33\%}& {70.93\%}\\ \hline

{DenseNet77} & {90.23\%} & {90.91\%}& {90.83\%}& {65.90\%} & {67.60\%}& {67.36\%}\\ \hline

{DenseNet100} & {91.40\%} & {91.70\%}& {91.90\%}& {67.80\%} & {67.80\%}& {67.89\%}\\ \hline

\hline
\end{tabular}
\end{table*}

\subsection{Model Training Stage}
\label{sec:training}
In order to investigate and understand the effect of various BatchNorm types on the model noise-resistant property, the models under consideration are designed with a BatchNorm layer at every layer. This design approach is chosen to ensure that the benefit of the BatchNorm layer on the training process is fully enabled. Specifically, a model with $L_2$ BatchNorm type is first designed and trained from scratch until convergence is achieved. The training process is repeated with the BatchNorm type changed to the $L_1$ and $TopK$ BatchNorm types and their performances compared.

\subsection{Model Inference stage}
\label{sec:inference}
In order to compare the effect of the various BatchNorm type on the noise-resistant property of the model, the model performance in the presence of noise is measured using an appropriate performance metric. The model classification accuracy(\%) is used as the performance metric as the models are classification models, and the analog noise used in this paper is modeled as an additive Gaussian white noise.

The white Gaussian additive noise, which is injected into the model weights of zero mean and a standard deviation of $\sigma_{noise}$, is used in this work. The value of $\sigma_{noise}$, defined as the energy/power of the noise, is obtained by benchmarking it against the standard deviation of the weight as shown in equation (\ref{equ1}). Benchmarking it against the standard deviation of the weight ensures that the weights are injected with a proportional noise. This is done by using the appropriate signal to noise ratio ($SNR$) or noise form factor (${\eta}$) values as stated in equation (\ref{equ2}) and (\ref{equ3}), respectively. In this work, Gaussian noise with zero mean with standard deviation with a form factor ($\eta$) values of 1\%, 4\%, 8\%, 12\%, 16\% and 20\% is injected into the model weights to determine the model noise-resistant property.
\begin{equation}
\label{equ1}
{\sigma_{noise}}=\frac{\sigma_w}{SNR}
\end{equation}
\begin{equation}
\label{equ2}
{\eta}=\frac{1}{SNR}
\end{equation}

\begin{equation}
\label{equ3}
{\sigma_{noise}}={\eta}\times{\sigma_w}
\end{equation}
\newcommand{\comment}[1]{}

The model with $L_2$ BatchNorm type obtained in Section \ref{sec:training} are all put in inference mode, and the value of the model test classification accuracy of the model is evaluated using the test dataset to establish the baseline inference performance. The baseline classification test accuracy is equivalent to injecting additive analog Gaussian white noise of zero mean and standard deviation of noise form factor ($\eta$) of 0\%.

After that, an additive analog Gaussian noise of zero mean and standard deviation equal to the form factor $\eta$ of x\% of the $\sigma_w$ at a layer $i$ is injected into all the model weights. The standard deviation value is equivalent to adding noise with power equivalent to $SNR$ value of $\frac{100}{x}$. The new model inference accuracy due to the injected noise is then evaluated on the test dataset. The steps above are then repeated multiple times with the $\eta$ value of x\% in order to obtain average inference classification accuracy as shown in equation (\ref{avg}). The average classification accuracy due to the presence of the noise is then normalized with the baseline inference classification accuracy using the formulae in equation (\ref{equ4}).

\begin{equation}
\label{avg}
{a_{avr}}=\frac{\sum\limits_{i=1}^{N} a_i}{N}
\end{equation}

\begin{equation}
\label{equ4}
{A^x}_{}=\frac{{a^x_{avg}}} {{a}_o}
\end{equation}
where ${a}_o$ are the baseline classification accuracy, ${A^x}_{}$ and ${a^x_{avg}}$ are the normalized classification accuracy and average classification accuracy due to the present of noise of $\eta$ value of $x\%$ respectively. The procedure above is then repeated with model with $L_1$ BatchNorm and $TopK$ BatchNorm types and the values of the normalized classification inference accuracy are compared and analyzed.

\subsection{Software and Hardware}
\label{subsec:ExpSetup}
All the models used in this work are trained and tested using PyTorch deep learning framework installed on the NVIDIA V100-DGXS-32GB GPU.

\section{RESULTS and ANALYSIS}
\label{sec:results}

\begin{figure*}[htbp]
\centering
\includegraphics[width=18cm]{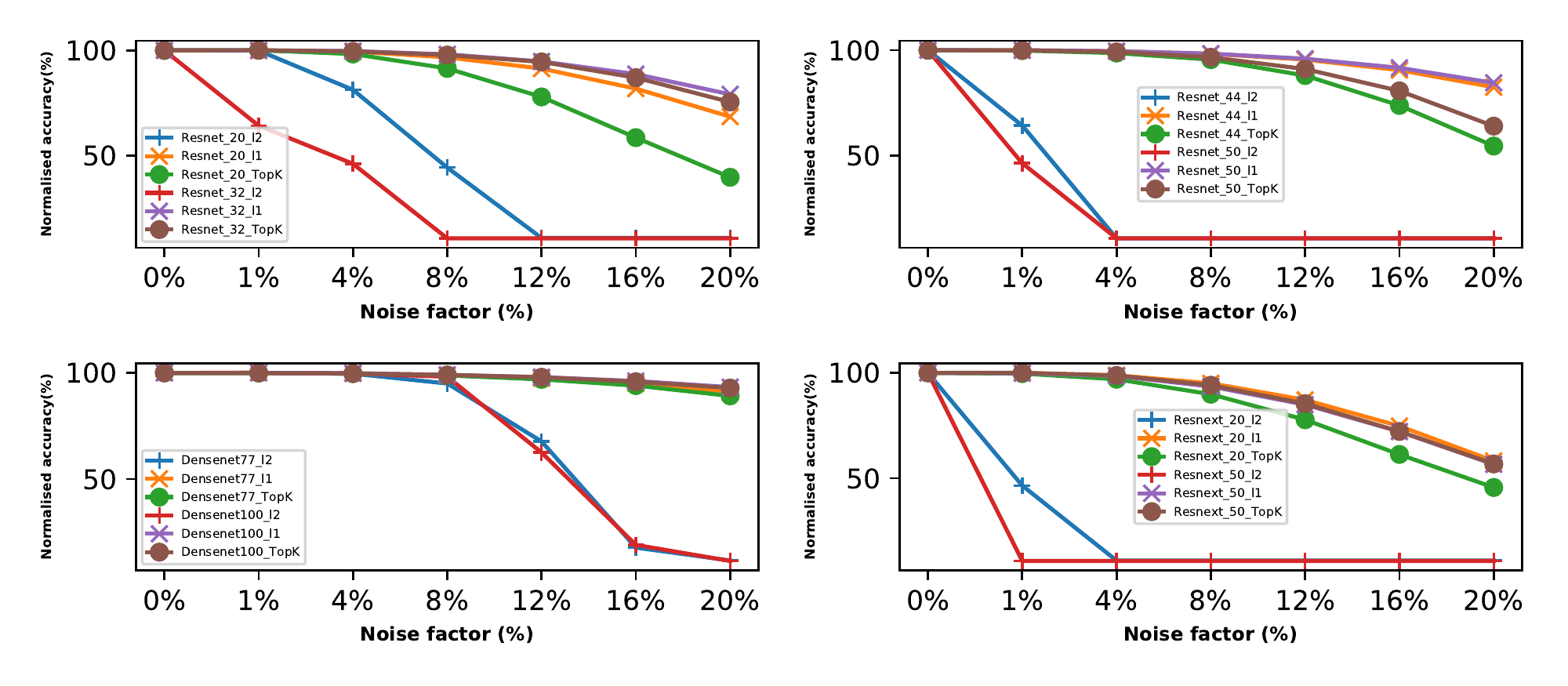}
\caption{The comparison of the  inference performance of ResNet, ResNext and DenseNet models with $L_2$, $L_1$, and $TopK$ BatchNorm types trained on CIFAR10 dataset when Gaussian noise is added to all the model weights. The performance metric is the normalized inference accuracy}
\label{cifar10NormalisedImage}
\end{figure*}

\begin{figure*}[htbp]
\centering
\includegraphics[width=18cm]{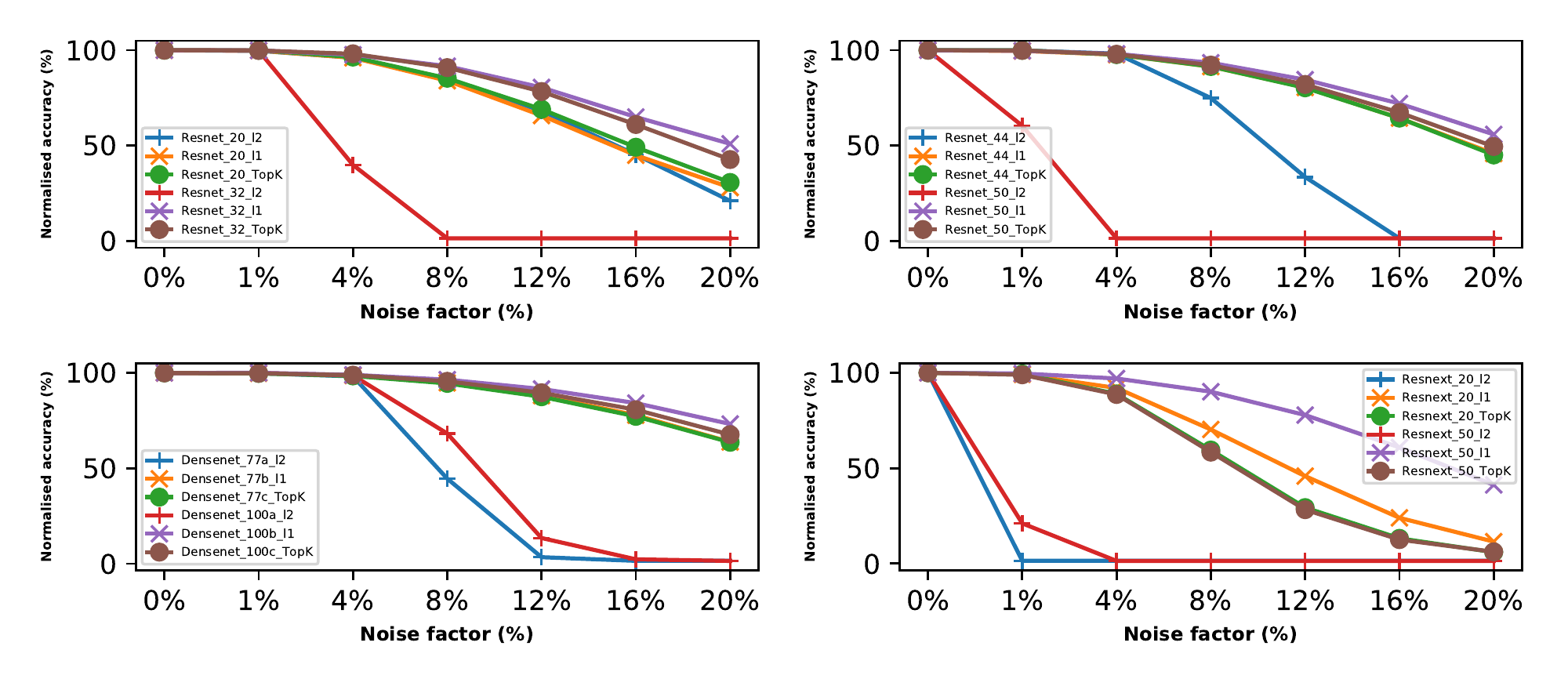}
\caption{The comparison of the  inference performance of ResNet, ResNext and DenseNet models with $L_2$, $L_1$, and $TopK$ BatchNorm types trained on CIFAR100 dataset when Gaussian noise is added to all the model weights. The performance metric is the normalized inference accuracy}
\label{cifar100NormalisedImage}
\end{figure*}

\begin{table*}
\centering
\caption{Comparison of the average normalized percentage classification accuracy of ResNet, ResNext and DenseNet model with various BatchNorm Types over 5 non-baseline noise factor in the presence of gaussian noise in all its layer during inference when tested with CIFAR10 and CIFAR100 dataset. The performance metric is the average normalized classification accuracy as defined in equation (\ref{equu}).}
\label{tab:result_cifar100-cifar10}
\begin{tabular}{|c|c|c|c|c|c|c|c|c|c|c|c|c|c|c|}
\hline
\multirow{3}{4em}{} & \multicolumn{6}{|c|}{Dataset}\\ \hline

& \multicolumn{3}{|c|}{CIFAR10}& \multicolumn{3}{|c|}{CIFAR100}\\ \hline

{Model Name} & {$L_2$ BN} & {$L_1$ BN} & {TopK BN}& {$L_2$ BN} & {$L_1$ BN} & {TopK BN}\\ \hline

{Resnet18} & {43.10\%} & {89.64\%}& {77.68\%}& {69.20\%} & {69.80\%}& {71.79\%}\\ \hline

{Resnet32} & {25.63\%} & {93.38\%}& {92.41\%}& {24.25\%} & {80.95\%}& {78.52\%}\\ \hline

{Resnet44} & {19.68\%} & {94.39\%}& {85.07\%}& {51.58\%} & {79.92\%}& {79.83\%}\\ \hline

{Resnet56} & {16.72\%} & {95.01\%}& {88.60\%}& {11.25\%} & {83.96\%}& {81.49\%}\\ \hline

{ResNext18} & {16.85\%} & {85.66\%}& {78.52\%}& {1.46\%} & {57.139\%}& {49.40\%}\\ \hline

{Resnext56} & {10.74\%} & {84.26\%}& {84.55\%}& {4.67\%} & {77.80\%}& {48.83\%}\\ \hline

{DenseNet77} & {65.27\%} & {97.05\%}& {96.43\%}& {41.52\%} & {87.18\%}& {86.84\%}\\ \hline

{DenseNet100} & {65.15\%} & {97.74\%}& {97.56\%}& {47.37\%} & {90.76\%}& {88.71\%}\\ \hline

\hline
\end{tabular}
\end{table*}

This section presents and discusses the results obtained from the experiment in Section \ref{sec:methods} by comparing the noise-resistant property of a model with $L_2$, $L_1$ and $TopK$ batch normalization types. The result is achieved by injecting gaussian noise into the model weights, causing a weight change, and measuring the performance due to the weight change. The performance metric is the inference accuracy as the model under consideration are classification models.

\subsubsection{Baseline Performance}
\label{sec:baseline}
The baseline inference accuracy of the various model under consideration with all the batch normalization types are presented in Table \ref{tab:result_baseline} for CIFAR10 and CIFAR100 datasets. The baseline accuracy is the inference accuracy of the model obtained just after training. No analog noise has been injected into the weights of the model, which is equivalent to a model injected with Gaussian noise of zero mean and standard deviation of 0. The baseline inference accuracy of all the models trained on the CIFAR10 dataset is greater than the equivalent model trained on the CIFAR100 dataset. This is expected as the classification task on the CIFAR100 dataset is more complex than that on the CIFAR10 dataset.
Furthermore, models with $L_2$ BatchNorm have the most model with the highest performance across the two datasets, and this explains why $L_2$ BatchNorm type is the most popular BatchNorm type. However, the performance difference between the models with $L_2$ BatchNorm and other BatchNorm types is very marginal. In fact, there are models where the variant with the best model inference baseline accuracy is the model with $L_1$ BatchNorm (DenseNet77 and DenseNet100 models). Hence, these results make a case for the use of other BatchNorm types apart from $L_2$ BatchNorm as the use of other types does not necessarily come at a considerable performance cost.

\subsubsection{Performance in the presence of Analog noise}
\label{sec:noise_performance}

The results obtained by injecting Gaussian noise of various noise factors into the weight of models with different BatchNorm types are shown in Figures \ref{cifar10NormalisedImage} and \ref{cifar100NormalisedImage}. In these Figures, the normalized inference accuracy (defined in equation (\ref{avg})) is plotted against the noise form factor (defined in equation (\ref{equ2})) of the noise injected into the model for various classification tasks. The noise form factor of $0\%$ is equivalent to the baseline model without any noise present, which explains why the normalized average inference accuracy is 100\%.

In general, It is observed that the model normalized inference accuracy decreases with increase in the analog noise power level across all models and datasets. For example, $L_1$ variant of ResNet18 model trained on CIFAR10 dataset has a normalized inference accuracy of 99.94\%, 99.45\%, 96.70\%, 91.35\%, 81.85\%, and 68.55\% when injected with noise with $\eta$ value of 1\%, 4\%, 8\%, 12\%, 16\%, and 20\%, respectively. The $TopK$ variant of ResNet18 model trained on CIFAR10 dataset also has a normalized inference accuracy value of 100\%, 98.27\%, 91.25\%, 77.92\%, 58.60\%, and 39.79\% when injected with noise with $\eta$ value of 1\%, 4\%, 8\%, 12\%, 16\%, and 20\%, respectively.
This observation is in agreement with the trend observed by the authors in~\cite{Fagbohungbe2020BenchmarkingIP} as the ability of the noise to corrupt the model weight increases with the noise power level. At a low noise power level, It is observed that the model suffers little or no reduction in the model normalized inference accuracy, and at elevated noise power level, the model inference accuracy does not fail suddenly but gradually. This behavior speaks to the inherent noise-resistant property of models as noted in \cite{BatchNorm2020,Merolla2016DeepNN,bay1}. However, the rate of decrease of the model normalized average inference accuracy varies from model to model and task to task for a fixed noise power level.

\subsubsection{Impact of BatchNorm Type on Model Noise Resistant Property}
\label{sec:BatchNorm}

Figures \ref{cifar10NormalisedImage} and \ref{cifar100NormalisedImage} also shows that models with $L_2$ BatchNorm types suffer more degradation or reduction in model normalized inference accuracy when compared to same model with $L_1$ and TopK BatchNorm types. From Figure \ref{cifar10NormalisedImage}, It can be observed that the normalized accuracy of $L_2$, $L_1$, and $TopK$ variants of DenseNet100 model trained on CIFAR10 datasets when injected with gaussian noise of zero mean and standard deviation of noise factor of 20\% is 11.64\%, 90.90\% and 89.18\%, respectively. Similarly, the normalized inference accuracy of $L_2$, $L_1$, and $TopK$ variants of DenseNet100 model trained on CIFAR100 datasets when injected with gaussian noise of noise factor level of 20\% is 1.47\%, 73.24\% and 67.66\%, respectively. This observation is valid for all other models under consideration in this paper.

In order to aid better summarization and analysis of results in Figures \ref{cifar10NormalisedImage} and \ref{cifar100NormalisedImage}, a new metric is introduced which is mathematically defined below:

\begin{equation}
\label{equu}
A_{avr}=\frac{\sum\limits_{i=1}^{N} A^x_i}{N}
\end{equation}
where $A_{avr}$ is the average normalized percentage classification accuracy and $N$ is the number of non-baseline noise factor which is 6. This new metric summarizes the performance of all the model trained on the CIFAR10 and CIFAR100 datasets over the 6 non-baseline noise factor and the result is given in Table \ref{tab:result_cifar100-cifar10}.
From Table \ref{tab:result_cifar100-cifar10}, It can be observed that for a particular model, the model variant with $L_2$ BatchNorm type has the lowest average normalized inference accuracy. For example, for the CIFAR10 dataset, the average normalized inference accuracy for ResNet56 with $L_2$ BatchNorm is 16.72\%, which is lower than 95.1\% and 88.60\% which are the values of average normalized inference accuracy for the same model with $L_1$ BatchNorm and $TopK$ BatchNorm types respectively. This observation also holds true for various BatchNorm type variants of ResNet56 model trained CIFAR100 dataset with an average normalized inference accuracy of 11.25\%, 83.96\% and 81.49\% for $L_2$, $L_1$ and $TopK$ BatchNorm variants of ResNet56 respectively. These observations also hold for the other models under consideration in this work.
It is instructive to note that the $L_2$ BatchNorm variant of all models trained performed poorly compared to the $L_1$ and $TopK$ variant of all the models. This poor performance means that the noise-resistant property of these models can be remarkably improved by changing the BatchNorm type from $L_2$ type to $L_1$ or $TopK$ type. This proposal is because these models' $L_1$ or $TopK$  BatchNorm variants have a higher average normalized inference accuracy value than the $L_2$ BatchNorm variant. In addition, these change in the BatchNorm type does not come with any significant performance cost as the baseline of the inference accuracy of the models $L_1$ and $TopK$ variant are very close to the values of the $L_2$ variant. In addition, it is easier to perform the computation of $L_1$ and $TopK$ BatchNorm on hardware platforms than $L_2$ BatchNorm.

\subsubsection{Impact of Model Architecture and Type on Model Noise Resistant Property}
\label{sec:BatchNorm}

It can be observed that value of average normalized inference accuracy value differs significantly across rows and columns in Table \ref{tab:result_cifar100-cifar10}. This is very instructive when combined with Figures \ref{cifar10NormalisedImage} and \ref{cifar100NormalisedImage}. For a fixed BatchNorm type variant of all models trained on CIFAR10 datatype, the value of average normalized accuracy differs significantly, and this observation is also true for these models trained on CIFAR100 datatype. For example, the $TopK$ BatchNorm variant of ResNet18, ResNet32, ResNet44, ResNet56, and ResNext56 are 77.68\%, 92.41\%, 85.07\%, 88.06\%, and 84.55\%, respectively. Hence, the noise-resistant property of a model is influenced by the model architecture. Also, the BatchNorm types are also model architecture features, and the analysis shows that the value of the average normalized inference accuracy of a particular model for different BatchNorm types differs.
Furthermore, the noise-resistant ability of a model with a specific architecture differs when trained on a different dataset. For example, the average normalized inference accuracy of DenseNet77 model with $L_1$ BatchNorm is trained on CIFAR10 and CIFAR100 is 97.05\% and 87.18\%, respectively. Similarly, the average normalized inference accuracy for $TopK$ BatchNorm variant of the DenseNet77 model trained on CIFAR10 and CIFAR100 datasets are 96.43\% and 86.84\%, respectively. It can be safely concluded that the noise-resistant ability of a model as measured by the average normalized inference accuracy of the model is reduced as the complexity of the dataset the model is trained on increases. This observation is in agreement with the observations made by authors in \cite{BatchNorm2020}.

\section{Related Works and Discussions}
\label{sec:discussion}

The $L_2$ BatchNorm is introduced to accelerate and stabilize model training of deep models by reducing the internal covariate shift in layers or subnetwork of a DNN model~\cite{Ioffe2015BatchNA}. Although the work in~\cite{Santurkar2018HowDB} agrees that $L_2$ BatchNorm is very effective as stated in \cite{Ioffe2015BatchNA}, it contends that the effectiveness of $L_2$ BatchNorm is due to its ability to smoothing the optimization landscape significantly and enables training with a higher learning rate. In order to solve the poor performance of $L_2$ BatchNorm when training size is small and when samples are not independent, various variants such as batch renormalization~\cite{Ioffe2017BatchRT}, group normalization~\cite{Wu2018GroupN}, weight normalization~\cite{Salimans2016WeightNA} has been introduced to solve this problem. Dues to its popularity, the effect of $L_2$ BatchNorm on the noise-resistant property of DNN models, is investigated in \cite{BatchNorm2020}.

The need to deploy DNN models on analog hardware, despite its noisy nature, has led to the introduction of methods of improving the noise-resistant property of DNN models. One of the popular methods introduced is the noise injection method~\cite{Fagbohungbe2020BenchmarkingIP,joshi}, which involves adding noise to the weight or gradient of the model during training. The authors in \cite{NoisyNN} demonstrated that even more improved noise-resistant property can be obtained when the noise injection method is combined with knowledge distillation. Training models in a noisy hardware environment similar to the ones they will be deployed in order to condition them is also proposed in~\cite{Bo,schmid}. The impact of learning rate value on model noise-resistant property is studied in \cite{bay1}. The use of linear and non-linear analog error correction codes to protect weights and bias of the DNN model is demonstrated in \cite{Upadhyaya2019ErrorCF,Upadhyaya2019ErrorCF2}. An algorithm based on deep reinforcement learning, which uses a selection protection scheme to choose a critical bit for error correction code (ECC) protection, is developed in~\cite{huang2020functional}. The work in~\cite{Zhang2019} introduces the use of a generalized fault-aware pruning technique to improve model resilience to hardware fault. Lastly, the work in~\cite{ShieldeNN} introduces a framework that synergies the mitigation approach and computational resources using a neural network architecture-aware partial replacement approach to identify the parameters of interest in the consecutive network layers for hardening to improve model resilience.

This paper differs from existing work because it does not use a noise injection method or error correction code to improve DNN model noise-resistant property. It investigates the noise-resistant property of building blocks of DNN models and then uses the obtained information to design models that are noise resistant. In this case, the use of $L_1$ and $TopK$ BatchNorm type is proposed to design a noise-resistant model for analog hardware ahead of the very popular $L_2$ BatchNorm. This conclusion is made after investigating the noise-resistant property of $L_2$, $L_1$, and $TopK$ BatchNorm types using different model architecture. This approach can be combined with other methods to provide an extra layer of protection against noise. This paper is closely related to the work in~\cite{BatchNorm2020} although this work is different as it proffers a solution to the negative impact of $L_2$ BatchNorm. Although noise in this paper is modeled as additive Gaussian noise as done in~\cite{joshi,NoisyNN}, this paper differs in that it does not use noise to improve its noise-resistant property as done in these papers. This work is also different from~\cite{Qiao2019MicroBatchTW,QinBitFlipping} where the noise is modeled as digital noise. Furthermore, it is also different from \cite{Santurkar2018HowDB,Bjorck2018UnderstandingBN} which aims to provide alternate reasons and mathematical derivation why batch normalization accelerates training.

\section{{Conclusion}}
\label{sec:conclusion}

The use of $L_1$ and $TopK$ BatchNorm types for training DNN models for analog accelerators is proposed and investigated in this work. The justification for the use of the proposed BatchNorm type is done by comparing the performance of $L_1$ and $TopK$ BatchNorm variant of DNN models with the $L_2$ BatchNorm variant of the same DNN models in the absence and presence of analog noise. The noise in this study is modeled as additive Gaussian noise.

The result of this study established that DNN models with $L_2$ BatchNorm have poor noise-resistant property as observed in \cite{BatchNorm2020}. This study also showed that models with $L_1$ and $TopK$ BatchNorm type have excellent noise-resistant property when noise is injected into all the weights in the model. This is established by comparing the normalized inference accuracy due to noise on the various BatchNorm type variants of the models. Furthermore, the performance of these $L_1$ and $TopK$ BatchNorm type model variants is competitive with the performance of the $L_2$ BatchNorm variant in the absence of noise. This comparison, done using the inference accuracy of the models, demonstrated that changing from the $L_2$ BatchNorm type does not come at a significant inference performance cost. Finally, it is also observed that the complexity of the classification type also affects the model noise-resistant property irrespective of the BatchNorm type being used.


\section{Acknowledgments}
\label{acknowledgement}

This research work is supported in part by the U.S. Army Research Office (ARO) award W911NF-2110143 and the U.S. Office of the Under Secretary of Defense for Research and Engineering (OUSD(R\&E)) under agreement number FA8750-15-2-0119. The U.S. Government is authorized to reproduce and distribute reprints for governmental purposes notwithstanding any copyright notation thereon. The views and conclusions contained herein are those of the authors and should not be interpreted as necessarily representing the official policies or endorsements, either expressed or implied, of the ARO or the Office of the Under Secretary of Defense for Research and Engineering (OUSD(R\&E)) or the U.S. Government.

\bibliographystyle{IEEEtran}

\bibliography{BatchNormalisation}

\begin{thebibliography}{10}
\providecommand{\url}[1]{#1}
\csname url@samestyle\endcsname
\providecommand{\newblock}{\relax}
\providecommand{\bibinfo}[2]{#2}
\providecommand{\BIBentrySTDinterwordspacing}{\spaceskip=0pt\relax}
\providecommand{\BIBentryALTinterwordstretchfactor}{4}
\providecommand{\BIBentryALTinterwordspacing}{\spaceskip=\fontdimen2\font plus
\BIBentryALTinterwordstretchfactor\fontdimen3\font minus
  \fontdimen4\font\relax}
\providecommand{\BIBforeignlanguage}[2]{{%
\expandafter\ifx\csname l@#1\endcsname\relax
\typeout{** WARNING: IEEEtran.bst: No hyphenation pattern has been}%
\typeout{** loaded for the language `#1'. Using the pattern for}%
\typeout{** the default language instead.}%
\else
\language=\csname l@#1\endcsname
\fi
#2}}
\providecommand{\BIBdecl}{\relax}
\BIBdecl

\bibitem{resnet}
\BIBentryALTinterwordspacing
K.~He, X.~Zhang, S.~Ren, and J.~Sun, ``Deep residual learning for image
  recognition,'' \emph{CoRR}, vol. abs/1512.03385, 2015. [Online]. Available:
  \url{http://arxiv.org/abs/1512.03385}
\BIBentrySTDinterwordspacing

\bibitem{Burrgeo2021}
G.~W. Burr, A.~Sebastian, T.~Ando, and W.~Haensch, ``Ohm's law + kirchhoff's
  current law = better ai: Neural-network processing done in memory with analog
  circuits will save energy,'' \emph{IEEE Spectrum}, vol.~58, no.~12, pp.
  44--49, 2021.

\bibitem{Onasami2022UnderwaterAC}
O.~Onasami, D.~Adesina, and L.~Qian, ``Underwater acoustic communication
  channel modeling using deep learning,'' 2022.

\bibitem{Yang2022AJE}
B.~Yang, O.~Fagbohungbe, X.~Cao, C.~Yuen, L.~Qian, D.~T. Niyato, and Y.~Zhang,
  ``A joint energy and latency framework for transfer learning over 5g
  industrial edge networks,'' \emph{IEEE Transactions on Industrial
  Informatics}, vol.~18, pp. 531--541, 2022.

\bibitem{Fagbohungbe2021EfficientPP}
O.~Fagbohungbe, S.~R. Reza, X.~Dong, and L.~Qian, ``Efficient privacy
  preserving edge intelligent computing framework for image classification in
  iot,'' \emph{IEEE Transactions on Emerging Topics in Computational
  Intelligence}, 2021.

\bibitem{Kariyappa}
S.~Kariyappa, H.~Tsai, K.~Spoon, S.~Ambrogio, P.~Narayanan, C.~Mackin, A.~Chen,
  M.~Qureshi, and G.~W. Burr, ``Noise-resilient dnn: Tolerating noise in
  pcm-based ai accelerators via noise-aware training,'' \emph{IEEE Transactions
  on Electron Devices}, vol.~68, no.~9, pp. 4356--4362, 2021.

\bibitem{mixedsignal}
\BIBentryALTinterwordspacing
M.~Klachko, M.~R. Mahmoodi, and D.~B. Strukov, ``Improving noise tolerance of
  mixed-signal neural networks,'' \emph{CoRR}, vol. abs / 1904.01705, 2019.
  [Online]. Available: \url{http://arxiv.org/abs/1904.01705}
\BIBentrySTDinterwordspacing

\bibitem{Li}
H.~{Li}, M.~{Bhargav}, P.~N. {Whatmough}, and H.~. {Philip Wong}, ``On-chip
  memory technology design space explorations for mobile deep neural network
  accelerators,'' in \emph{2019 56th ACM/IEEE Design Automation Conference
  (DAC)}, 2019, pp. 1--6.

\bibitem{joshi}
\BIBentryALTinterwordspacing
V.~Joshi, M.~L. Gallo, I.~Boybat, S.~Haefeli, C.~Piveteau, M.~Dazzi,
  B.~Rajendran, A.~Sebastian, and E.~Eleftheriou, ``Accurate deep neural
  network inference using computational phase-change memory,'' \emph{CoRR},
  vol. abs / 1906.03138, 2019. [Online]. Available:
  \url{http://arxiv.org/abs/1906.03138}
\BIBentrySTDinterwordspacing

\bibitem{charan}
G.~{Charan}, A.~{Mohanty}, X.~{Du}, G.~{Krishnan}, R.~V. {Joshi}, and Y.~{Cao},
  ``Accurate inference with inaccurate rram devices: A joint algorithm-design
  solution,'' \emph{IEEE Journal on Exploratory Solid-State Computational
  Devices and Circuits}, vol.~6, no.~1, pp. 27--35, 2020.

\bibitem{NoisyNN}
C.~Zhou, P.~Kadambi, M.~Mattina, and P.~N. Whatmough, ``Noisy machines:
  Understanding noisy neural networks and enhancing robustness to analog
  hardware errors using distillation,'' \emph{ArXiv}, vol. abs/2001.04974,
  2020.

\bibitem{han2015deep}
S.~Han, H.~Mao, and W.~J. Dally, ``Deep compression: Compressing deep neural
  networks with pruning, trained quantization and huffman coding,'' \emph{arXiv
  preprint arXiv:1510.00149}, 2015.

\bibitem{luo2017thinet}
J.-H. Luo, J.~Wu, and W.~Lin, ``Thinet: A filter level pruning method for deep
  neural network compression,'' in \emph{Proceedings of the IEEE international
  conference on computer vision}, 2017, pp. 5058--5066.

\bibitem{hinton2015distilling}
G.~Hinton, O.~Vinyals, and J.~Dean, ``Distilling the knowledge in a neural
  network,'' \emph{arXiv preprint arXiv:1503.02531}, 2015.

\bibitem{tpu}
\BIBentryALTinterwordspacing
N.~P. Jouppi, C.~Young, N.~Patil, D.~Patterson, G.~Agrawal, R.~Bajwa, S.~Bates,
  S.~Bhatia, N.~Boden, A.~Borchers, R.~Boyle, P.-l. Cantin, C.~Chao, C.~Clark,
  J.~Coriell, M.~Daley, M.~Dau, J.~Dean, B.~Gelb, T.~V. Ghaemmaghami,
  R.~Gottipati, W.~Gulland, R.~Hagmann, C.~R. Ho, D.~Hogberg, J.~Hu, R.~Hundt,
  D.~Hurt, J.~Ibarz, A.~Jaffey, A.~Jaworski, A.~Kaplan, H.~Khaitan,
  D.~Killebrew, A.~Koch, N.~Kumar, S.~Lacy, J.~Laudon, J.~Law, D.~Le, C.~Leary,
  Z.~Liu, K.~Lucke, A.~Lundin, G.~MacKean, A.~Maggiore, M.~Mahony, K.~Miller,
  R.~Nagarajan, R.~Narayanaswami, R.~Ni, K.~Nix, T.~Norrie, M.~Omernick,
  N.~Penukonda, A.~Phelps, J.~Ross, M.~Ross, A.~Salek, E.~Samadiani, C.~Severn,
  G.~Sizikov, M.~Snelham, J.~Souter, D.~Steinberg, A.~Swing, M.~Tan,
  G.~Thorson, B.~Tian, H.~Toma, E.~Tuttle, V.~Vasudevan, R.~Walter, W.~Wang,
  E.~Wilcox, and D.~H. Yoon, ``In-datacenter performance analysis of a tensor
  processing unit,'' \emph{SIGARCH Comput. Archit. News}, vol.~45, no.~2, p.
  1–12, jun 2017. [Online]. Available:
  \url{https://doi.org/10.1145/3140659.3080246}
\BIBentrySTDinterwordspacing

\bibitem{gpu}
A.~Maghazeh, U.~D. Bordoloi, P.~Eles, and Z.~Peng, ``General purpose computing
  on low-power embedded gpus: Has it come of age?'' in \emph{2013 International
  Conference on Embedded Computer Systems: Architectures, Modeling, and
  Simulation (SAMOS)}, 2013, pp. 1--10.

\bibitem{Chen}
A.~Chen, S.~Ambrogio, P.~Narayanan, H.~Tsai, C.~Mackin, K.~Spoon, A.~Friz,
  A.~Fasoli, and G.~W. Burr, ``Enabling high-performance dnn inference
  accelerators using non-volatile analog memory (invited),'' in \emph{2020 4th
  IEEE Electron Devices Technology Manufacturing Conference (EDTM)}, 2020, pp.
  1--4.

\bibitem{xiao}
T.~P. Xiao, C.~H. Bennett, B.~Feinberg, S.~Agarwal, and M.~J. Marinella,
  ``Analog architectures for neural network acceleration based on non-volatile
  memory,'' \emph{Applied Physics Reviews}, vol.~7, no.~3, 9 2020.

\bibitem{Chang}
H.-Y. Chang, P.~Narayanan, S.~C. Lewis, N.~C.~P. Farinha, K.~Hosokawa,
  C.~Mackin, H.~Tsai, S.~Ambrogio, A.~Chen, and G.~W. Burr, ``Ai hardware
  acceleration with analog memory: Microarchitectures for low energy at high
  speed,'' \emph{IBM Journal of Research and Development}, vol.~63, no.~6, pp.
  8:1--8:14, 2019.

\bibitem{Kornilios2020}
\BIBentryALTinterwordspacing
K.~Kourtis, M.~Dazzi, N.~Ioannou, T.~Grosser, A.~Sebastian, and E.~Eleftheriou,
  ``Compiling neural networks for a computational memory accelerator,''
  \emph{CoRR}, vol. abs/2003.04293, 2020. [Online]. Available:
  \url{https://arxiv.org/abs/2003.04293}
\BIBentrySTDinterwordspacing

\bibitem{Chutengarmibm}
\BIBentryALTinterwordspacing
C.~Zhou, F.~Garc{\'{\i}}a{-}Redondo, J.~B{\"{u}}chel, I.~Boybat, X.~T. Comas,
  S.~R. Nandakumar, S.~Das, A.~Sebastian, M.~L. Gallo, and P.~N. Whatmough,
  ``Analognets: {ML-HW} co-design of noise-robust tinyml models and always-on
  analog compute-in-memory accelerator,'' \emph{CoRR}, vol. abs/2111.06503,
  2021. [Online]. Available: \url{https://arxiv.org/abs/2111.06503}
\BIBentrySTDinterwordspacing

\bibitem{Fagbohungbe2020BenchmarkingIP}
O.~I. Fagbohungbe and L.~Qian, ``Benchmarking inference performance of deep
  learning models on analog devices,'' \emph{ArXiv}, vol. abs/2011.11840, 2020.

\bibitem{Merolla2016DeepNN}
P.~Merolla, R.~Appuswamy, J.~Arthur, S.~K. Esser, and D.~Modha, ``Deep neural
  networks are robust to weight binarization and other non-linear
  distortions,'' \emph{ArXiv}, vol. abs/1606.01981, 2016.

\bibitem{Santurkar2018HowDB}
S.~Santurkar, D.~Tsipras, A.~Ilyas, and A.~Madry, ``How does batch
  normalization help optimization?'' in \emph{NeurIPS}, 2018.

\bibitem{Bjorck2018UnderstandingBN}
J.~Bjorck, C.~P. Gomes, and B.~Selman, ``Understanding batch normalization,''
  in \emph{NeurIPS}, 2018.

\bibitem{Ioffe2017BatchRT}
S.~Ioffe, ``Batch renormalization: Towards reducing minibatch dependence in
  batch-normalized models,'' in \emph{NIPS}, 2017.

\bibitem{Ioffe2015BatchNA}
S.~Ioffe and C.~Szegedy, ``Batch normalization: Accelerating deep network
  training by reducing internal covariate shift,'' \emph{ArXiv}, vol.
  abs/1502.03167, 2015.

\bibitem{Wu2019L1B}
S.~Wu, G.~Li, L.~Deng, L.~Liu, D.~Wu, Y.~Xie, and L.~Shi, ``\$l1\$ -norm batch
  normalization for efficient training of deep neural networks,'' \emph{IEEE
  Transactions on Neural Networks and Learning Systems}, vol.~30, pp.
  2043--2051, 2019.

\bibitem{BatchNorm2020}
O.~Fagbohungbe and L.~Qian, ``Effect of batch normalization on noise resistant
  property of deep learning models,'' 2021.

\bibitem{xie2017aggregated}
S.~Xie, R.~Girshick, P.~Dollár, Z.~Tu, and K.~He, ``Aggregated residual
  transformations for deep neural networks,'' 2017.

\bibitem{huang2018densely}
G.~Huang, Z.~Liu, L.~van~der Maaten, and K.~Q. Weinberger, ``Densely connected
  convolutional networks,'' 2018.

\bibitem{bay1}
O.~Fagbohungbe and L.~Qian, ``Impact of learning rate on noise resistant
  property of deep learning models,'' 2022.

\bibitem{Wu2018GroupN}
Y.~Wu and K.~He, ``Group normalization,'' in \emph{ECCV}, 2018.

\bibitem{Salimans2016WeightNA}
T.~Salimans and D.~P. Kingma, ``Weight normalization: A simple
  reparameterization to accelerate training of deep neural networks,'' in
  \emph{NIPS}, 2016.

\bibitem{Bo}
G.~M. {Bo}, D.~D. {Caviglia}, and M.~{Valle}, ``An on-chip learning neural
  network,'' in \emph{Proceedings of the IEEE-INNS-ENNS International Joint
  Conference on Neural Networks. IJCNN 2000. Neural Computing: New Challenges
  and Perspectives for the New Millennium}, vol.~4, 2000, pp. 66--71 vol.4.

\bibitem{schmid}
A.~Schmid, Y.~Leblebici, and D.~Mlynek, ``Mixed analogue-digital
  artificial-neural-network architecture with on-chip learning,''
  \emph{Circuits, Devices and Systems, IEE Proceedings -}, vol. 146, pp. 345 --
  349, 01 2000.

\bibitem{Upadhyaya2019ErrorCF}
P.~Upadhyaya, X.~Yu, J.~Mink, J.~Cordero, P.~Parmar, and A.~Jiang, ``Error
  correction for hardware-implemented deep neural networks,'' 2019.

\bibitem{Upadhyaya2019ErrorCF2}
P.~Upadhyaya, X.~Yu, J.~Mink, C.~J., P.~Parmar, and A.~Jiang, ``Error
  correction for noisy neural networks,'' 10 2019.

\bibitem{huang2020functional}
K.~Huang, P.~Siegel, and A.~Jiang, ``Functional error correction for robust
  neural networks,'' 2020.

\bibitem{Zhang2019}
J.~J. Zhang, K.~Basu, and S.~Garg, ``Fault-tolerant systolic array based
  accelerators for deep neural network execution,'' \emph{IEEE Design Test},
  vol.~36, no.~5, pp. 44--53, 2019.

\bibitem{ShieldeNN}
N.~Khoshavi, A.~Roohi, C.~Broyles, S.~Sargolzaei, Y.~Bi, and D.~Z. Pan,
  ``Shieldenn: Online accelerated framework for fault-tolerant deep neural
  network architectures,'' in \emph{2020 57th ACM/IEEE Design Automation
  Conference (DAC)}, 2020, pp. 1--6.

\bibitem{Qiao2019MicroBatchTW}
S.~Qiao, H.~Wang, C.~Liu, W.~Shen, and A.~Yuille, ``Micro-batch training with
  batch-channel normalization and weight standardization,'' 2019.

\bibitem{QinBitFlipping}
M.~Qin, C.~Sun, and D.~Vucinic, ``Improving robustness of neural networks
  against bit flipping errors during inference,'' \emph{Journal of Image and
  Graphics}, vol.~6, pp. 181--186, 01 2018.

\end{thebibliography}

$ $ \\

\end{document}